# Physics-Informed Graph Neural Network for Spatial-temporal Production Forecasting

Wendi Liu[*,1], Michael J. Pyrcz[1,2]

**Abstract**

Production forecast based on historical data provides essential value for developing hydrocarbon resources. Classic history matching workflow is often computationally intense and geometry-dependent. Analytical data-driven models like decline curve analysis (DCA) and capacitance resistance models (CRM) provide a grid-free solution with a relatively simple model capable of integrating some degree of physics constraints. However, the analytical solution may ignore subsurface geometries and is appropriate only for specific flow regimes and otherwise may violate physics conditions resulting in degraded model prediction accuracy. Machine learning-based predictive model for time series provides non-parametric, assumption-free solutions for production forecasting, but are prone to model overfit due to training data sparsity; therefore may be accurate over short prediction time intervals.

We propose a grid-free, physics-informed graph neural network (PI-GNN) for production forecasting. A customized graph convolution layer aggregates neighborhood information from historical data and has the flexibility to integrate domain expertise into the data-driven model. The proposed method relaxes the dependence on close-form solutions like CRM and honors the given physics-based constraints. Our proposed method is robust, with improved performance and model interpretability relative to the conventional CRM and GNN baseline without physics constraints.

**Introduction**

Production forecasting plays an essential role in reserve estimation and supports decision-making for subsurface resource development. Widely applied physics-informed predictive models vary in complexity from simple, grid-free methods like decline curve analysis (DCA) or capacitance resistance models (CRM), to more complicated grid-based history-matching based on reservoir model and numerical simulation. DCA is an empirical regression model based on historical production data and remains the cornerstone for production forecasting due to its simplicity and fast computation. The parameter estimation for the DCA model relies on assumptions of certain physics conditions. Arps' (1945) DCA method is empirically derived from historical data. Fetkovich et al. (1996) justified the use of Arps' parameters assuming pseudo-steady state flow (PSS) and constant pressure at the inner boundary. Fetkovich (1980) relaxes the PSS assumption by combining two families of type curves, encompassing both transient and PSS flow periods while retaining a constant flowing bottomhole pressure (BHP) constraint. Rate-transient analysis (RTA) introduces both variable rate and BHP by history-matching the rate-time data with an appropriate type curve (Palacio and Blasingame, 1993; Mishra, 2014), as opposed to pressure-transient analysis that requires a long period of well shut-in. RTA can further extend to multi-well formulation unlike traditional DCA (Marhaendrajana and Blasingame, 2001). However, RTA does require reservoir properties to estimate a constant drainage area for well-by-well analysis while CRM can estimate drainage volumes only using BHP and flow rate, though still assuming constant drainage volume.

---

[*] Corresponding author wendi_liu@utexas.edu
[1] Hildebrand Department of Petroleum and Geosystem Engineering, Cockrell School of Engineering, The University of Texas at Austin, USA
[2] Department of Geological Sciences, Jackson School of Geosciences, The University of Texas at Austin, USA

CRM comprises a family of material balance system without the requirement of a grid-based geological model and assists geological analysis by providing inter-well connectivity estimation based on historical data only (Holanda et al., 2018). CRM was initially developed for water flooding to optimize injection rate allocation (Yousef et al., 2006; Sayarpour et al., 2009). The governing differential equation of CRM with producer-based control volume representation for water flooding is

$$C_t V_p \frac{d\bar{p}(t)}{dt} = I(t) - q(t) \qquad (1)$$

The linear productivity model is

$$q(t) = J[\bar{p}(t) - p_{wf}(t)] \qquad (2)$$

Replacing $\bar{p}$ term, average pressure within drainage volume, of Eq. 1 with Eq. 2, results in

$$\tau \frac{dq(t)}{dt} + q(t) = FI(t) - \frac{dp_{wf}(t)}{dt} J \cdot \tau \qquad (3)$$

where $\tau = \frac{C_t V_p}{J}$, $C_t$ is the total compressibility, $V_p$ is the drainage volume of a producer, $J$ is the productivity index, $p_{wf}$ is the well bottomhole pressure, $q$ is the production rate, $I$ is the injector rate and $F$ is the connectivity matrix between injector and producer. For CRM, the estimated production rate of well $j$ at $t_n$ timestep is an analytical integration assuming linear variation of BHP and stepwise changes in injection rate derived from Eq. 3,

$$q_j(t_n) = q_j(t_0)e^{-\frac{t_n-t_0}{\tau_j}} + \sum_{k=1}^{n}\left[e^{-\frac{t_n-t_k}{\tau_j}}\left(1-e^{-\frac{\Delta t_k}{\tau_j}}\right)\left(\sum_{i}^{N_I} F_{ij}I_i^{(k)} - \tau_j \cdot J_j \frac{\Delta p_{wf,j}^{(k)}}{\Delta t_k}\right)\right] \quad (4)$$

where $I_i^{(k)}$ represents the injection rate of $i$th injector and $\Delta p_{wf,j}^{(k)}$ is the changes of $j$th producer BHP in the $k$th time interval $t_k - t_{k-1}$. CRM can be further extended to primary recovery to help determine reservoir compartmentalization (Nguyen et al., 2011; Izgec and Kabir, 2012) and to tertiary recovery (Laochamroonvorapongse et al., 2014).

The above predictive models are derived based on historical data and physics-informed assumptions that may not hold for complex pressure conditions, for example, long-period transient flow like in shale reservoirs. Machine learning methods have been widely used to accommodate production forecasting in those situations due to their ability to provide non-parametric, assumption-free, non-linear function fitting. Recurrent neural networks (RNN), e.g., long short-term memory (LSTM), gated recurrent units (GRU), have been recently applied to replace DCA for tight reservoir production predictions, motivated by their ability to capture time-dependence of the historical production data (Lee et al., 2019; Al-Shabandar et al., 2021). However, solely data-driven methods like deep learning lack interpretability and are difficult for domain experts to interpret and diagnose. Moreover, given the production data are sparsely sampled and the limited training data availability relative to the reservoir volume of interest, deep learning models with a large number of parameters are prone to overfit. Practical consequences of overfit machine learning forecasting models include degraded prediction accuracy beyond short intervals forward in time and poor ability to generalize to inform future exploration, motivating a merger of data-driven and physics-based models.

There is various recent research on leveraging physics information constraints to assist the deep learning training process and improve its predictive accuracy. Raissi et al. (2019) propose a physics-informed neural network (PINN) architecture that regularizes the training of deep neural networks to avoid overfitting and to generalize with sparse data by formulating non-linear partial differential equation (PDE) as part of the loss function and leveraging automatic differentiation of neural networks to solve the forward or inverse PDE problems, e.g., solve the Navier-Stokes equations for the velocity field and pressure distribution. The PINN architecture enables physics-informed neural network-based reservoir simulation surrogate models (Harp et al., 2021; Yang and Foster, 2021; Yan et al., 2022), which speeds up the traditional grid-based history matching workflow with deep learning and preserves the physics conditions for model fidelity. Physics constraints for grid-free analytical predictive models like CRM for water-flooding can also be incorporated into the loss function of multi-layer perceptron (MLP) models, i.e., a fully connected artificial neural network with multiple non-linear layers, to improve predictive accuracy for traditional CRM (Yewgat et al., 2020). Grid-free PINN models for production prediction are lightweight with no requirement for a geological model while the grid-based PINN reservoir simulation surrogate models are relatively more computationally intensive but have fewer simplifying assumptions to represent more complicated transport phenomena with complex geometry.

To benefit from grid-free modeling while honoring the complex topology and connectivity, of spatial-temporal datasets like production well network, graph neural network (GNN) models may be applied. Graphs are a type of data structure that models a set of objects (nodes or vertices) and their relationships (edges, i.e., the information of dependency between two nodes based on affinity, connectivity or adjacency) (Biggs et al., 1986). Graphs can describe a large number of systems across diverse areas with complicated geometry, topology, interconnectivity of objects, including social networks, molecular structures and spatial networks. Graph neural networks retain the expressive power and interpretability of graphs, as well as the flexibility to model any nonlinear function of neural networks. Similar to convolutional neural networks (CNN) or RNN preserving the information extracted from Euclidean data, i.e., grid-based data or 1D sequence, GNN generalizes the application of neural network to non-Euclidean geometry and provides inference of the given node with information aggregated from neighboring nodes. A schematic of CNN layer on a 2D grid-based data and GNN layer on a graph is shown in **Fig. 1**.

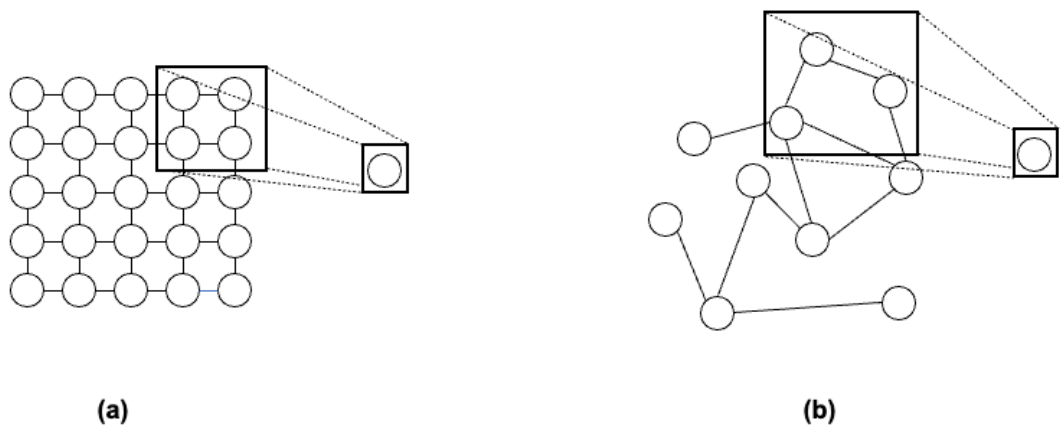

**Fig. 1** The schematic of (a) CNN layer on a 2D grid-based data and (b) GNN layer on a graph.

The variants of GNN can be categorized based on different aggregators to gather information from each node's neighbors and update the nodes' hidden states (Zhou et al., 2018). Two common aggregator types are convolution aggregator, e.g., Graph Convolutional Network (GCN, Kipf and Welling, 2017) and GraphSAGE (short for Graph Sample and Aggregate, Hamilton et al., 2017), and attention aggregator, e.g., Graph Attention Network (GAT, Velickovic et al., 2018). Convolution aggregators can be further separated into spectral methods and non-spectral (spatial methods) depending on how convolution works on graph representations. Spectral methods work with a spectral representation of the graphs, in which convolution operation is defined in the computation of eigen-decomposition of graph Laplacian, i.e., a matrix representation of a graph that summarizes the degree of connectivity of each node and the interconnectivity of the combinatorial of node pairs. One example of the spectral method is GCN, which has the layer-wise aggregation rule as

$$H^{(l+1)} = \sigma(\widehat{D}^{-\frac{1}{2}} \widehat{A} \widehat{D}^{-\frac{1}{2}} H^{(l)} W^{(l)}) \qquad (5)$$

Where $\widehat{A} = A + I_\mathcal{N}$ is the adjacency matrix (combinatorial of pairwise interconnectivity of nodes) of the undirected graph $\mathcal{G}$, $\widehat{A} \in \mathbb{R}^{\mathcal{N} \times \mathcal{N}}$. $I_\mathcal{N}$ is the identity matrix for graph self-connection. $\widehat{D}_{ii} = \sum_j \hat{A}_{ij}$ is the degree matrix. The normalized graph Laplacian is $L = I_\mathcal{N} - \widehat{D}^{-\frac{1}{2}} \widehat{A} \widehat{D}^{-\frac{1}{2}}$. $W^{(l)}$ denotes the trainable weight matrix of $l$th layer, i.e., a matrix of filter parameters in convolution. $\sigma(\cdot)$ is an activation function. $H^{(l+1)} \in \mathbb{R}^{\mathcal{N} \times \mathcal{D}}$ is the convolved signal matrix, where $\mathcal{N}$ is the number of nodes and $\mathcal{D}$ is the output dimension of the hidden state at $l+1$th layer. $H^{(0)} = X$, where $X$ is the matrix of node feature vectors $X_i$.

GNNs have been applied for spatial-temporal problems like traffic flow prediction by combining graph layers with RNN layers to achieve improved computational performance compared with RNN and better accuracy compared with classical temporal prediction statistical methods like autoregressive integrated moving average (ARIMA) (Yu et al., 2018; Pan et al., 2019; Ge et al., 2020). Similar frameworks have been extended to production forecasting problems. Wang et al. (2021) present a GCN-RNN-based framework as a data-driven reservoir surrogate model to estimate well connectivity for water-flooding production and predict BHP from historical data. The well connectivities between injectors and producers are informed by a trainable adjacency matrix. Although subsurface production data has similarities with traffic sensor data in terms of spatial-temporal dependency and complex topological relationships, the flow regime and production operations can further complicate the long-term accuracy of GNN-RNN model without physics constraints for production data. Therefore, it is necessary to incorporate the corresponding law of physics into the training process. Xing and Sela (2022) incorporate the physics laws in GNN training for state estimation of the water distribution system by penalizing the violation of macroscopic mass balance. However, in their method, the flow in the water pipeline assumes steady-state, i.e., the flow rate and pressure head are time-independent, which is very unlikely for reservoir flow regime. Another challenge is that traffic flow or water distribution systems already have a pre-defined topology structure, while the connection of well networks is implicit. The challenge remains to inform well graphs and integrate physics constraints for transport in porous media to provide a successful GNN model for production forecasting.

We propose a grid-free, physics-informed graph neural network (PI-GNN) method for subsurface production forecasting that is computationally efficient and incorporates the physics-based constraints, while providing better accuracy than traditional analytical grid-free data-driven models like CRM. The proposed model leverages the architecture of GNN to integrate domain expertise into the method by informing well adjacency matrix based on known geological interpretation. When domain expertise is not available or would benefit from data support, our proposed method has the option for a self-learnt adjacency matrix that quantifies and communicates well connectivity. We demonstrate the proposed method for waterflooding in a channel-like reservoir model with a variety of well patterns. Our proposed method achieves better performance compared with both classic CRM and the GNN model without physics constraints, while improving data and knowledge integration and model interpretability.

In the next section, we provide a detailed description of the proposed method, including the options to inform the graph adjacency matrix. In the results section, we demonstrate the proposed method with a 2D channel reservoir model for water flooding and compare the results with CRM and GNN models without physics constraints.

**Methodology**

This section consists of two major parts. The first part covers the calculation of the graph adjacency matrix for well topology, which can be informed either based on domain expertise or self-learnt during the training process. Next, we introduce the proposed model architecture, which is a customized neural network model composed of multiple functional blocks of MLP and GCN layers.

*Graph Construction*

Well connectivity graph is denoted as $\mathcal{G}(\mathcal{V}, \mathcal{E})$, where $\mathcal{V} = \{v_1, v_2, v_i ..., v_{N_I}; v_1, v_2, v_j ..., v_{N_P}\}$, representing $N_I$ injector nodes $v_i$ and $N_P$ producer nodes $v_j$, and $\mathcal{E} = \{e_{ij} | 1 \leq i \leq N_I, 1 \leq j \leq N_P\}$, indicating the connectivity between injectors and producers. The topology of well network is then described

in adjacency matrix as $\mathbf{A}_{N_I \times N_P} = (e_{ij})_{i \in N_I, j \in N_P}$. We suggest two options for informing adjacency matrix. One is based on domain expertise, e.g., available geological models, etc., which can be automated with the assistance of a workflow based on the fast marching method. Another way is initializing a trainable adjacency matrix $\mathbf{A}_{N_I \times N_P}$ of well connectivities for the proposed PI-GNN model and the connectivities are learnt while training the model. Graph adjacency matrix can be informed solely from training data in the second option, but with the sacrifice of a degree of user control. Users should consider the trade-off of professional time and user control when selecting between the two proposed methods for the calculation of the graph adjacency matrix.

Domain expertise can be integrated into the data-driven model by providing an estimated adjacency matrix that consists of binary matrix elements $e_{ij}$ denoting well connection as 1, and well no connection as 0. This estimated adjacency matrix works as prior knowledge to assist the PI-GNN model to learn well connectivity from the training data. To semi-automate the adjacency matrix estimation based on domain expertise, we suggest a workflow to assist the decision of the presence of connections between injectors and producers based on a threshold of travel time estimated from fast marching method given a geological model. Fast marching method is a computationally efficient method based on solving boundary values of the Eikonal equation using upwind finite-difference approximation (Sethian 1996). Eikonal equation may be expressed as

$$F(\mathbf{x})|\nabla t(\mathbf{x})| = 1, t \in \Omega \tag{6}$$

with boundary condition $t = g(\mathbf{x})$ on $\Gamma \in \Omega$, where $t(\mathbf{x})$ is the arrival time with grid system $\mathbf{x}$, $F(\mathbf{x})$ is the speed function and $\Gamma$ is any source location(s). The Eikonal equation can be applied to calculate pressure front propagation time based on reservoir properties and the specific Eikonal form derived from the pressure-diffusivity equation

$$\mu c_t(\mathbf{x})\phi(\mathbf{x})\frac{\partial p(\mathbf{x},t)}{\partial t} - \nabla \cdot [k(\mathbf{x})\nabla p(\mathbf{x},t)] = 0 \tag{7}$$

with certain assumptions, e.g., $\frac{\nabla k(\mathbf{x})}{k(\mathbf{x})}$ is negligible, where $k$ is the permeability, $\phi$ is the porosity, $C_t$ is the total compressibility, $\mu$ is the viscosity, $p$ is the pressure (Sharifi et al. 2014). The neighboring nodes of the graph are connected based on the shortest proxy pressure propagation time, i.e., earliest arrival, using the well location as the source with velocity field estimated from porosity, permeability and rock compressibility model assuming the viscosity and compressibility of the fluid do not change in the reservoir, i.e., $F(x) = \sqrt{\frac{k(x)}{\mu c_t(x)\phi(x)}}$. Quadrant search algorithm is implemented in the proposed workflow when searching for neighbor nodes to provide wide coverage of possible node connections. Quadrant search can also be replaced by octant search for a more elaborate search radially. In the workflow, the propagation time of each well location to the neighboring wells is calculated with fast marching method reimplemented in scikit-fmm (Furtney 2012).

**Algorithm 1**: Graph construction

1. for $i = 1,2 ... , N$:
2.     Calculate propagation time $t(x)$ with velocity field $F(x)$ with source $\Gamma_i$, which is $i$th injector location.
3.     for each quadrant / octant centered on injector $i$ at $\Gamma_i$:
4.         Find neighboring producer $j = 1, 2,...K$ locations with the least travel time $t(x_j)$, where $K$ is the number of producers to search within each quadrant/octant
5.         Connect the well indexes for the graph edge $e_{ij}$, i.e., $e_{ij} = 1$ in the adjacency matrix if wells are connected

*Model Construction*

Now we build a physics-informed graph neural network model (PI-GNN) customized for specific physics rules for the water flooding production forecasting. The architecture of the PI-GNN is summarized in **Fig. 2**.

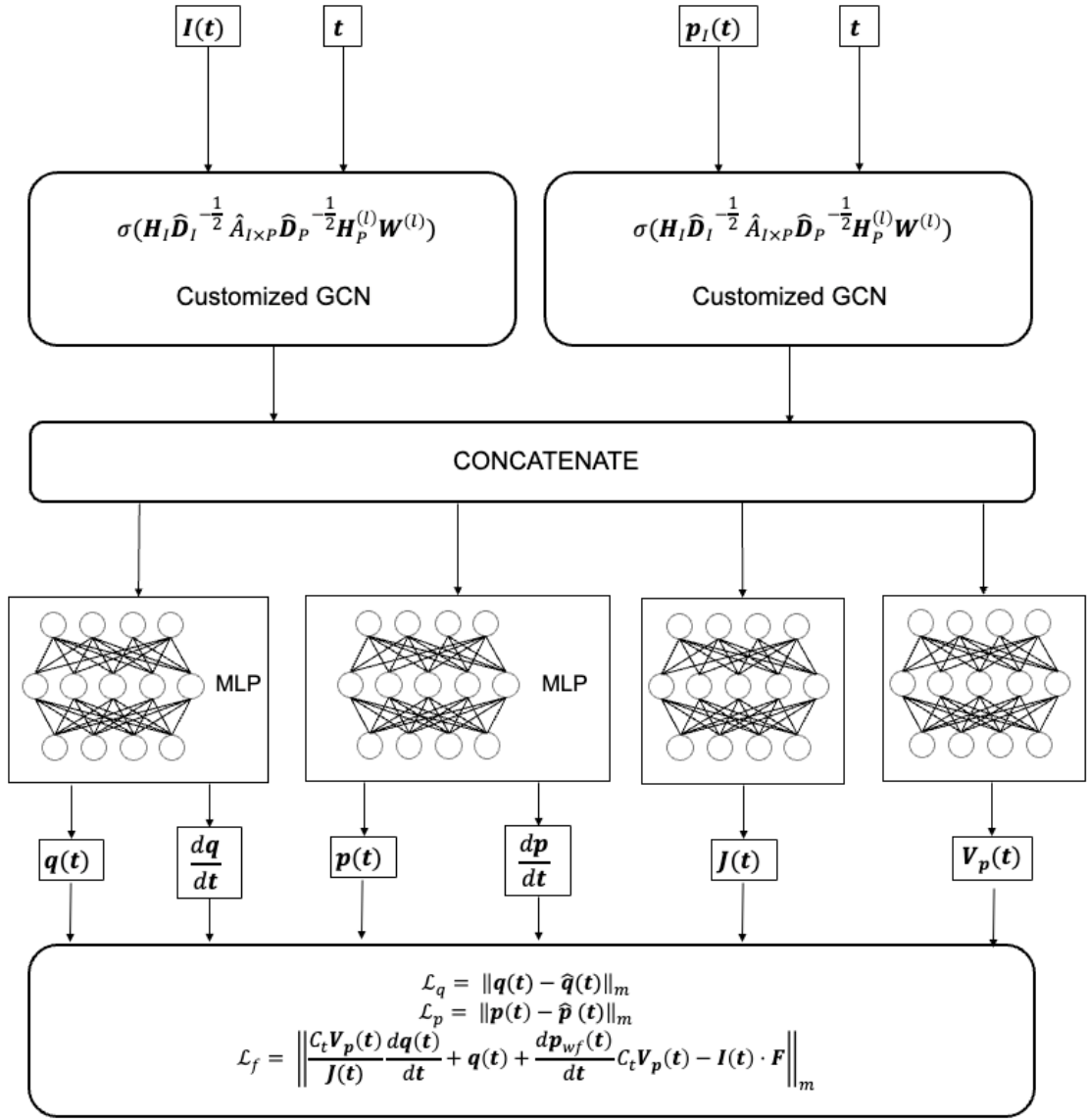

**Fig. 2** The schematic of PI-GNN architecture.

A customized graph convolutional layer aggregates the neighborhood information from injectors and combines with the current state of producers processing through a fully connected dense layer to update for the next hidden state of the next layer. The customized GCN layer is a variant of Eq. 5, where the adjacency matrix $A_{N_I \times N_P}$ can be either informed with the domain expertise to assist model convergence or initialize as a trainable model parameter with $\widehat{D}_{N_I}$ and $\widehat{D}_{N_P}$ representing the corresponding degree matrix for injectors and producers, respectively. The customized GCN layer is defined as

$$H^{(l+1)} = \sigma([H_{N_I}^{(l)} \widehat{D}_{N_I}^{-\frac{1}{2}} A_{N_I \times N_P} \widehat{D}_{N_P}^{-\frac{1}{2}} H_{N_P}^{(l)}] \cdot W^{(l)}) \tag{8}$$

where $H_{N_I}^{(l)}$ is the node predictor feature of injectors at $l$ layer, $H_{N_P}^{(l)}$ is the node predictor feature of producers, $H^{(l+1)}$ is the output hidden state at $l+1$ layer projected by graph Laplacian aggregating the features based on connectivity between injectors and producers. $W^{(l)}$ is the trainable weights at $l$ layer, similar to the $W^{(l)}$ in Eq. 5.

The PI-GNN model consists of 1 GNN block and 4 parallel MLP blocks. The GNN block contains two parallel customized GCN layers, where each GCN layer takes injector input, injection rate $I \in \mathbb{R}^{N_T \times N_I}$, and injector BHP $p_I \in \mathbb{R}^{N_T \times N_I}$, respectively. $N_T$ is the batch size for training the model, i.e., the length of the samples that pass through the model at one time. The parallel GCN layers have another common producer input, which is a sequence of number is formulated as timesteps $t$ for each producer, $t \in \mathbb{R}^{N_T \times N_P}$. All the inputs and outputs are scaled with a min-max scaler to avoid the impact of the imbalanced magnitude of the predictor features while preserving the shape of the original distributions. In the forward pass of the PI-GNN, the 4 MLP blocks output responses of production rate $q \in \mathbb{R}^{N_T \times N_P}$, producer BHP $p_{wf} \in \mathbb{R}^{N_T \times N_P}$, productivity index $J \in \mathbb{R}^{N_T \times N_P}$, and drainage volume $V_p \in \mathbb{R}^{N_T \times N_P}$, respectively. The corresponding loss function for this forward pass is also calculated. Among the 4 responses, production rate and producer BHP have observations in training data and are learnt in supervised manner with the corresponding loss function term penalizing the mismatch between predictions and observations. The productivity index and drainage volume training are "unsupervised" in terms of no available observations to calibrate predictions. The model learns the behavior of productivity index and drainage volume in an indirect way by penalizing the violation of the physics constraints of a customized loss function.

The physics constraints for the proposed PI-GCN are imposed in the customized loss function based on the material balance of CRM for water flooding while relaxing the assumptions required for obtaining the analytical solution for CRM. For CRM, the estimated production rate for each well in Eq. 4 is an analytical integration derived from Eq. 3 assuming linear variation of BHP, while PI-GNN does not require a closed-form function for estimation but finds the optimal parameters for non-linear neural network layers by minimizing the loss function. Also, CRM assumes linear productivity model with constant drainage volume $V_p$ and productivity index $J$ as parameters to estimate the production rate, while PI-GNN can estimate the time-dependent drainage volume $V_p(t)$ and productivity index $J(t)$ in the unsupervised manner as explained previously. The loss function of the PI-GNN consists of a mismatch term between observations and prediction for production rate, $\mathcal{L}_q$, mismatch term for producer BHP, $\mathcal{L}_q$, and physics constraint terms, $L_f$.

$$\mathcal{L}_q = \|q(t) - \hat{q}(t)\|_m \qquad (9)$$

$$\mathcal{L}_p = \|p(t) - \hat{p}(t)\|_m \qquad (10)$$

$$\mathcal{L}_f = \left\| \frac{C_t V_p(t)}{J(t)} \frac{dq(t)}{dt} + q(t) + \frac{dp_{wf}(t)}{dt} C_t V_p(t) - I(t) \cdot F \right\|_m \qquad (11)$$

where $\|\cdot\|_m$ is the $L^m$ norm function, specifically, $\mathcal{L}_q$ and $\mathcal{L}_p$ are the mean squared error (MSE) terms when $m = 2$. The physics constraint term $L_f$ is adapted from the CRM material balance with producer-based control volume in Eq. 3. $F \in \mathbb{R}^{N_I \times N_P}$ is the trainable well connectivity matrix initialized in the PI-GNN model as well as the trainable weight matrices $W$. When choosing self-learnt graph option for the model, $F$ is equivalent to adjacency matrix $A_{N_I \times N_P}$ for the customized GCN layer. For the domain-expertise-informed graph calculation option, a prior adjacency matrix is provided by the user for the GCN layer to assist well connectivity matrix $F$ training process. The $\frac{dq}{dt}$ and $\frac{dp_{wf}}{dt}$ terms in the loss function are calculated by automatic differentiation in the training process. The final loss function is optimized with Adam optimizer with gradient clip norm set to 1.0 to assist model convergence and gradient stability.

**Results**

To demonstrate our proposed PI-GNN method for production forecasting in a waterflooding, we build a synthetic channelized discrete facies reservoir truth model with complicated braided channel (net) facies connectivity within overbank (non-net) facies (Deutsch and Journel, 1997; Pyrcz et al., 2021). To demonstrate the robustness of the proposed PI-GNN model, we use the same tuned model hyperparameters for 4 different cases, which have the same injection schedule and the same producer well control in the same reservoir but with randomized well spatial locations. The locations of 2 injectors and 4 producers for the 4 cases with the reservoir permeability are shown in **Fig. 3**. The net, within channel permeability is 100 mD and non-net bank permeability is 1 mD. The reservoir porosity is constant at all locations, set to 15.0%. The production data are generated through reservoir simulation with the CMG IMEX black-oil simulator. The production time series data are split into training, validation and testing sets comprising 70%, 5% and 25% of the data, respectively. Training and validation sets are used for model training and tuning, in which validation loss is calculated so model training stops at the epoch with the least validation loss to constrain model complexity and to avoid model overfitting. Neural network models in general are stochastic due to the randomness in initialization and the stochastic optimization of loss function. Moreover, the proposed grid free PI-GNN model holds a higher degree of freedom as it has less constraint imposed from the number of independent variables ($t$) compared with other state-of-the-art grid-based PINN surrogate models ($x, y, z, t$), which results in a higher model variance with respect to the randomness. To mitigate the impact of randomness, we suggest using the model with the average of multiple realizations for improved model consistency. Therefore, the same training and tuning procedure is repeated for 10 different random seeds, and we use the average predictions of the 10 model realizations for evaluating the predictions over the withheld testing data.

We evaluate the PI-GNN model with a domain-expertise-informed adjacency matrix, where adjacency matrices inferred by domain expertise for the 4 cases are illustrated in **Fig. 5 (a)**, **Fig. 7(a)**, **Fig. 9(a)**, and **Fig. 11(a)** and also the corresponding trainable well connectivity matrix $F$ learnt from this model is shown in **Fig. 5 (b)**, **Fig. 7(b)**, **Fig. 9(b)**, and **Fig. 11(b)**. For reference, we also run the PI-GNN model with the trainable well connectivity matrix as adjacency matrix and the learnt adjacency matrices are shown in **Fig. 5 (c)**, **Fig. 7(c)**, **Fig. 9(c)**, and **Fig. 11(c)**. The two models provide similar well connectivity information while the PI-GNN with user-informed adjacency matrix speeds up model convergence compared with the self-learnt graph model.

We compare the predictions of PI-GNN with a class CRM baseline and a GNN model without physics constraints. The CRM baseline is calculated with the *CRMCompensated* function considering BHP in *pywaterflood* Python package (Male, 2022). The well connectivity matrices calculated from CRM for each case are shown in **Fig. 5 (d)**, **Fig. 7(d)**, **Fig. 9(d)**, and **Fig. 11(d)**. We build the GNN baseline model without physics constraints using the same customized GCN block with similar model complexity, i.e., the same magnitude of total trainable parameters, but without physics constraints. The corresponding self-learnt well connectivity matrices for each case are shown in **Fig. 5 (e)**, **Fig. 7(e)**, **Fig. 9(e)**, and **Fig. 11(e)**. The connectivity matrices from PI-GNN models and CRMs provide consistent information while the connectivity matrices from GNN baselines provide information that is against interpretation based on the truth model and inconsistent with the previous 3 models with physics constraints. The prediction results of the 4 models for case 1 through 4 are shown in **Fig. 4**, **Fig. 6**, **Fig. 8**, and **Fig. 10**, respectively. The left side of the vertical dash line is the predictions from the training process while the right side of the vertical dash line is the predictions for testing. From case 3, we can observe that our proposed method is robust, i.e., the domain-expertise-informed adjacency matrix does not need to be very accurate to converge to an improved performance and informative well connectivity.

To quantitatively evaluate the prediction performance, we use the root mean squared error (RMSE) as the evaluation metric

$$RMSE = \sqrt{\frac{\sum_{i=1}^{n}(y_{q_i}-\hat{y}_{q_i})^2}{n}} \quad (12)$$

where $y_{q_i}$ is the production rate observation in the testing dataset, $\hat{y}_{q_i}$ is the production rate prediction at the corresponding timestep, $i = 1,2,...,n$, $n$ is the number of the testing dataset. The testing RMSE of each producer and total testing RMSE for the models and baselines of 4 cases are shown in **Table 1**.

For this demonstration, our proposed PI-GNN method achieves the best prediction performance in terms of total testing RMSE compared with the two baseline methods. PI-GNN with a self-learnt graph has a higher model variance for different scenarios compared with PI-GNN with a domain-expertise-informed graph in terms of performance rank, which is logical considering the domain-expertise-informed adjacency matrix provides prior knowledge assisting model convergence and reducing model variance. The GNN method overall achieves improved accuracy compared with CRM as it does not predict based on closed-form solutions, which works for the scenarios when the analytical solution assumptions do not hold, e.g., the transient flow for certain producers before reaching pseudo-steady state when injectors abruptly change the injecting regimes. PI-GNN with self-learnt graph is able to learn the well connectivity that is more consistent with domain expertise and CRM compared with the GNN baseline without physics constraints. Physics constraints are especially essential for the data-driven models to learn informative well connectivity matrix judging from the consistency of well connectivity matrix from models with physics constraints, i.e., PI-GNN models and CRM, compared with GNN baselines.

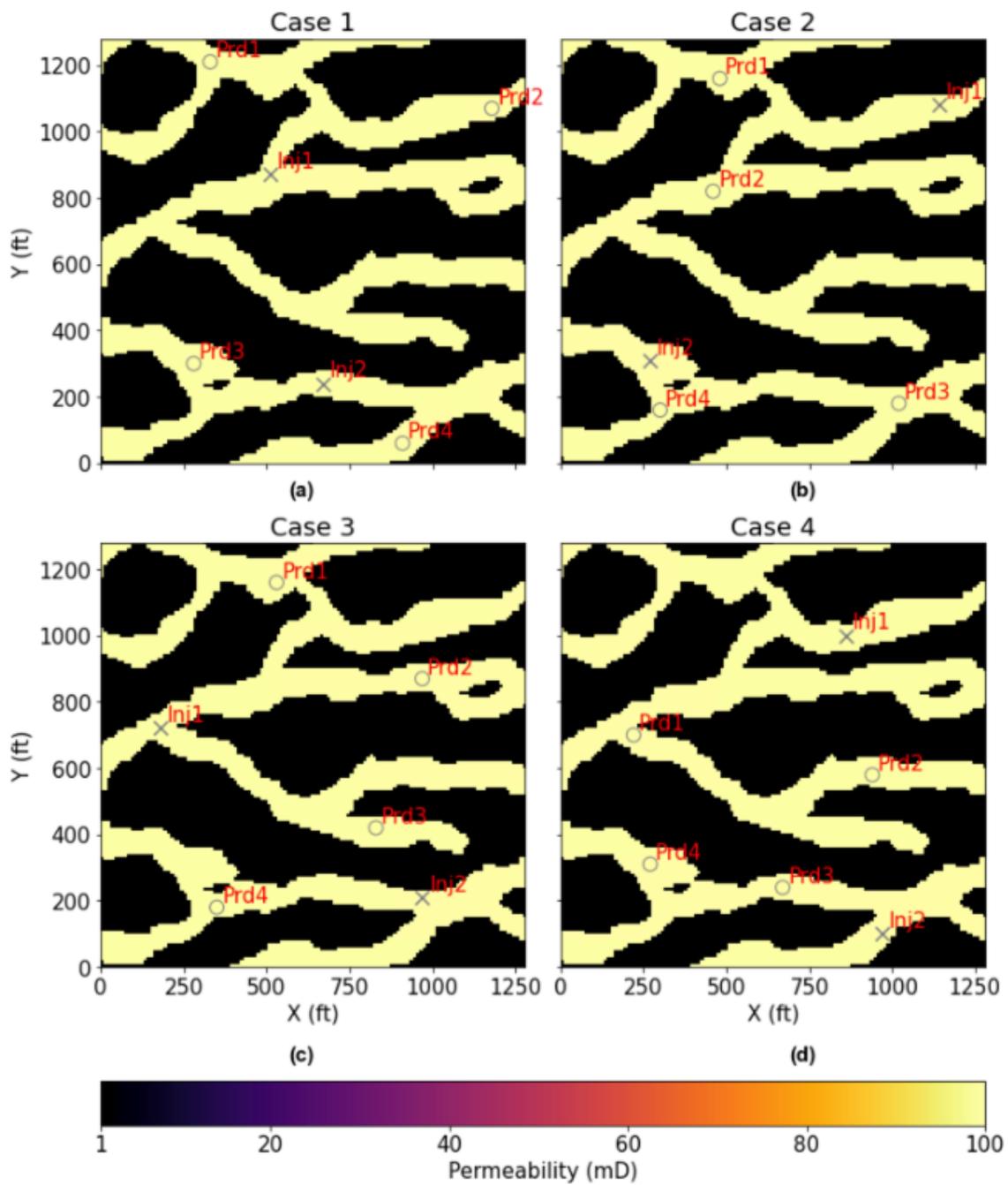

**Fig. 3** Reservoir permeability with injector (Inj) and producer (Prd) well locations for 4 cases, a – d.

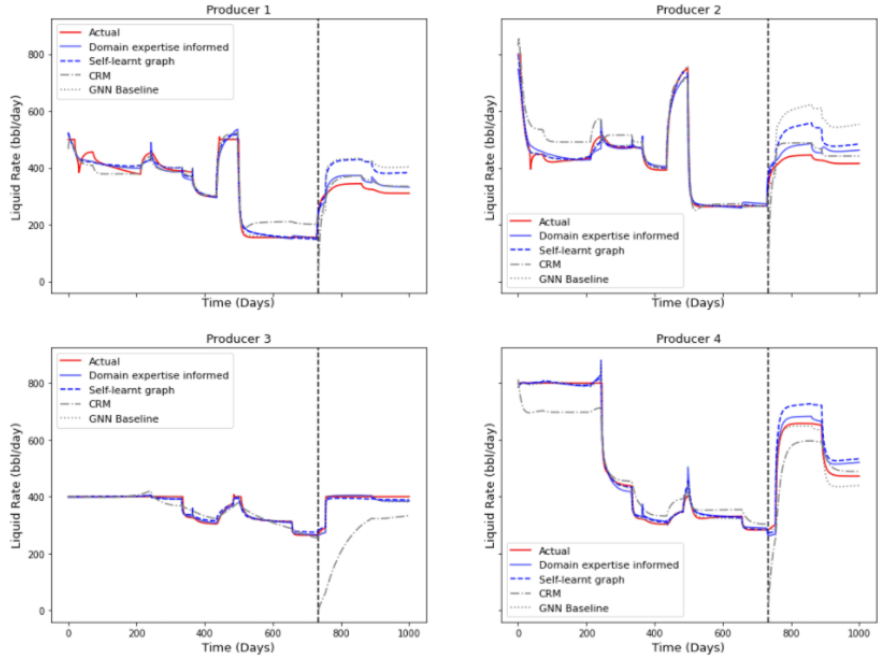

**Fig. 4** Case 1 production rate observations (red line) and predictions from PI-GNN model with user-informed adjacency matrix (solid line), PI-GNN with self-learnt adjacency matrix (dash line), CRM (dot-dash line), GNN without physics constraints (dot line). PI-GNN models are highlighted in blue.

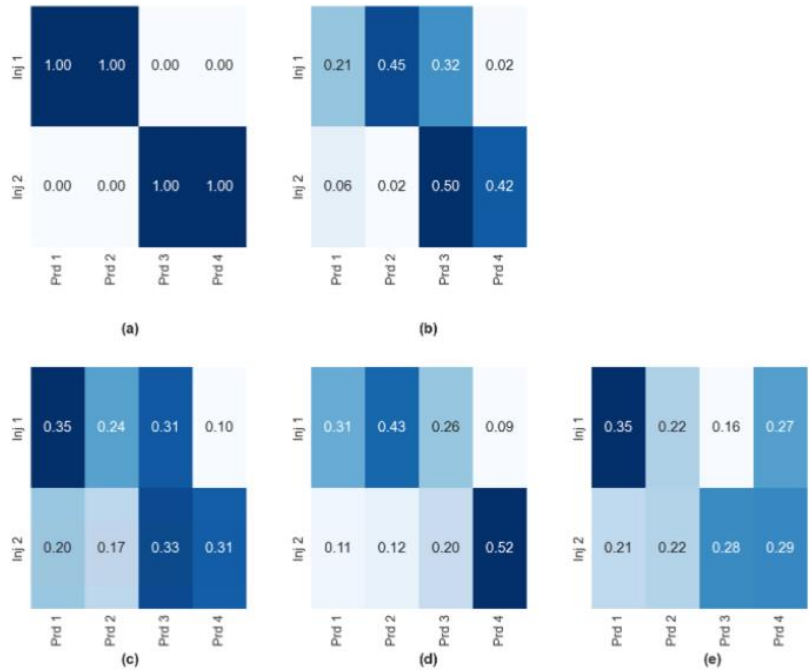

**Fig. 5** Case 1 (a) adjacency matrix informed by domain expertise (b) well connectivity matrix learnt from the PI-GNN model assisted with domain expertise (c) self-learnt well connectivity matrix with PI-GNN model (d) well connectivity matrix from CRM (e) well connectivity matrix leant from GNN baseline

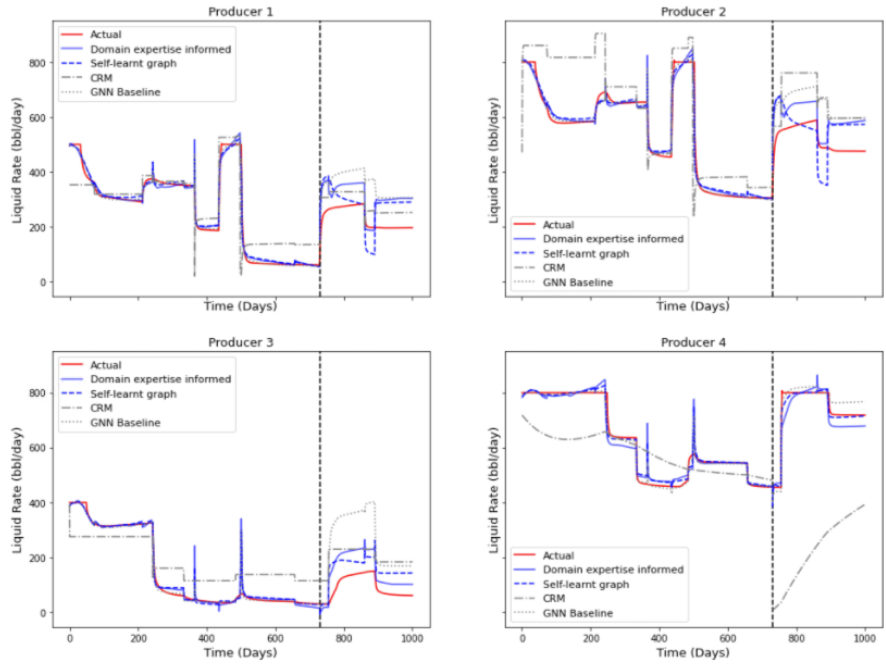

**Fig. 6** Case 2 production rate observations (red line) and predictions from PI-GNN model with user-informed adjacency matrix (solid line), PI-GNN with self-learnt adjacency matrix (dash line), CRM (dot-dash line), GNN without physics constraints (dot line). PI-GNN models are highlighted in blue.

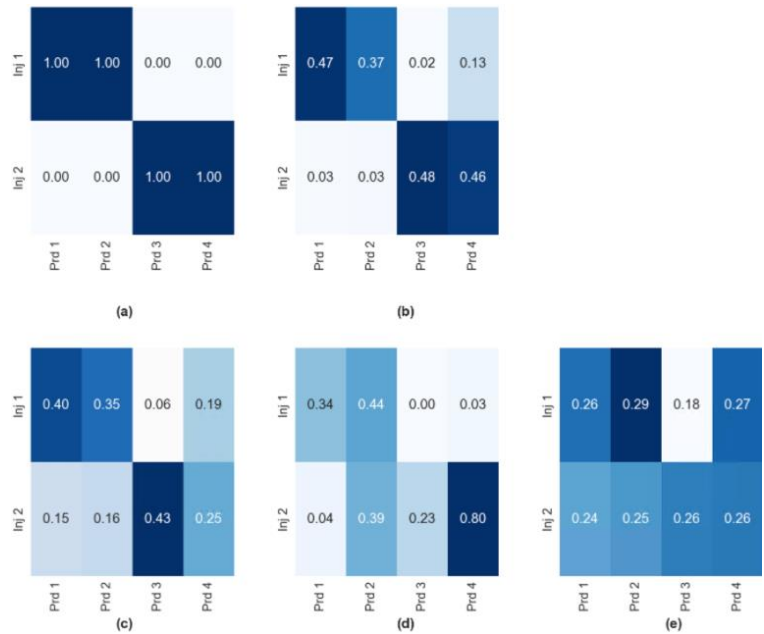

**Fig. 7** Case 2 (a) adjacency matrix informed by domain expertise (b) well connectivity matrix learnt from the PI-GNN model assisted with domain expertise (c) self-learnt well connectivity matrix with PI-GNN model (d) well connectivity matrix from CRM (e) well connectivity matrix leant from GNN baseline

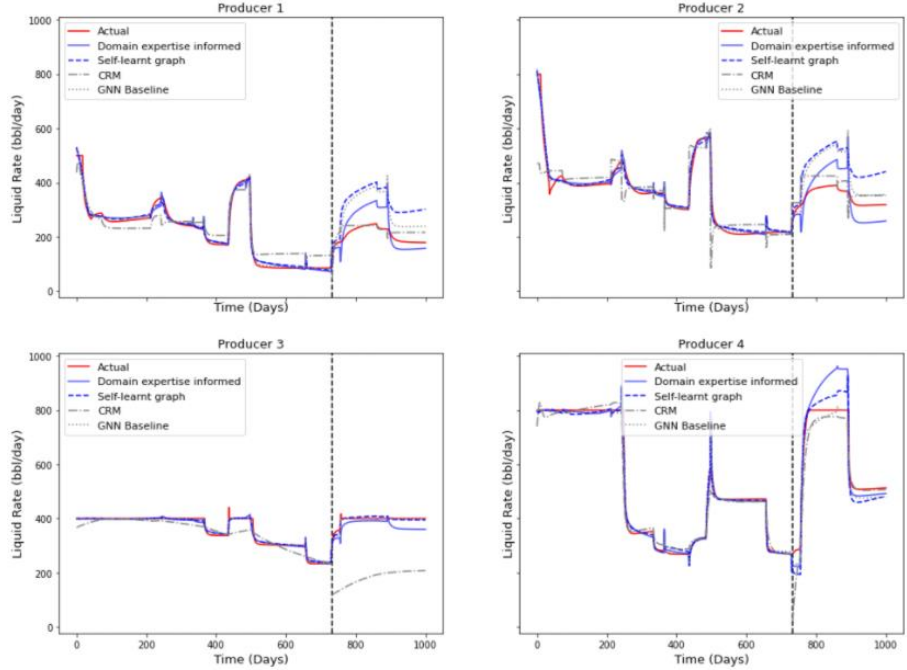

**Fig. 8** Case 3 production rate observations (red line) and predictions from PI-GNN model with user-informed adjacency matrix (solid line), PI-GNN with self-learnt adjacency matrix (dash line), CRM (dot-dash line), GNN without physics constraints (dot line). PI-GNN models are highlighted in blue.

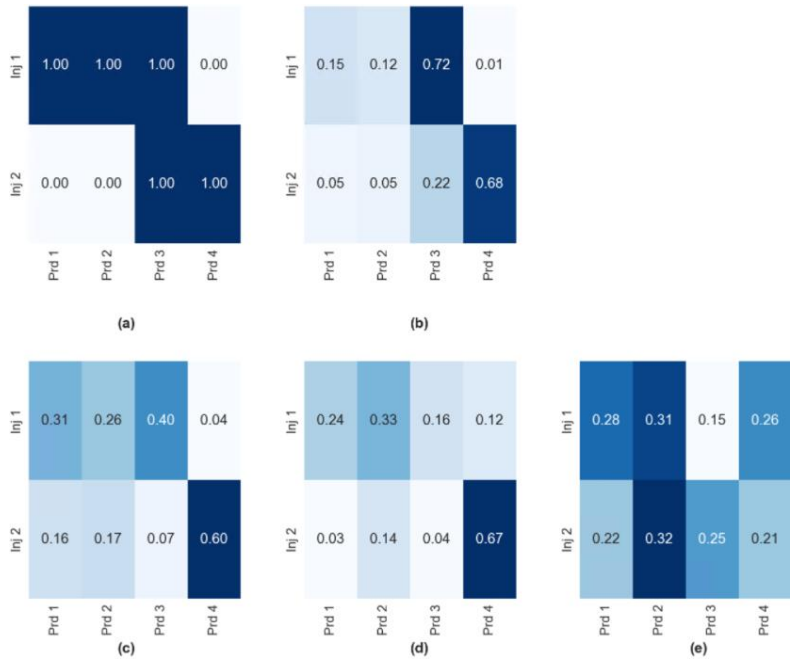

**Fig. 9** Case 3 (a) adjacency matrix informed by domain expertise (b) well connectivity matrix learnt from the PI-GNN model assisted with domain expertise (c) self-learnt well connectivity matrix with PI-GNN model (d) well connectivity matrix from CRM (e) well connectivity matrix leant from GNN baseline

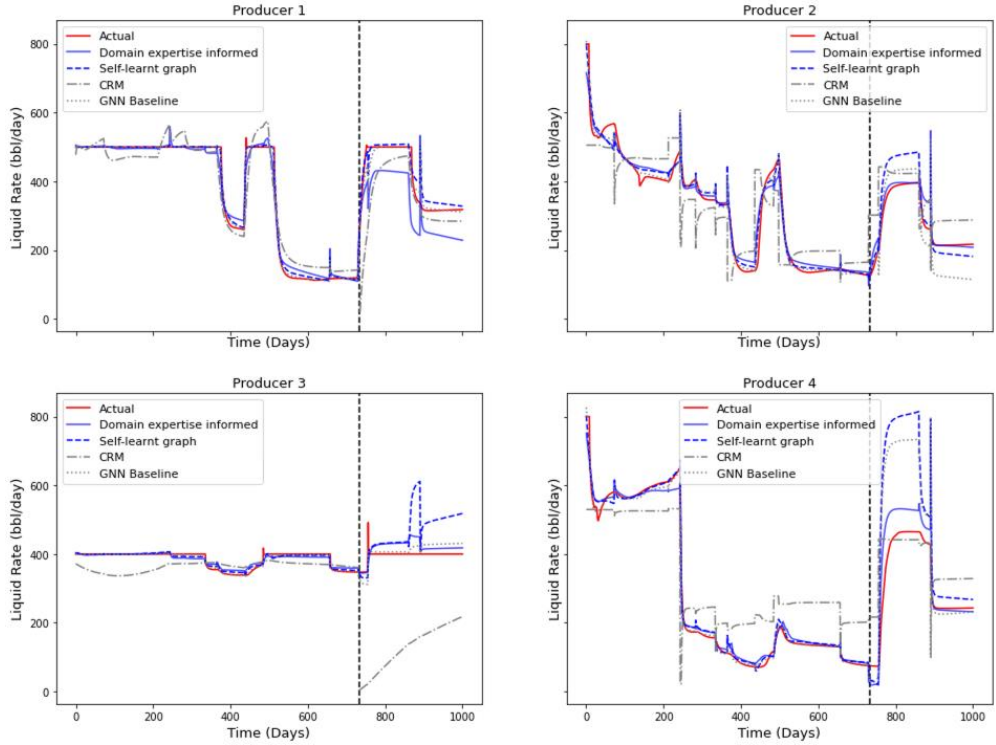

**Fig. 10** Case 4 production rate observations (red line) and predictions from PI-GNN model with user-informed adjacency matrix (solid line), PI-GNN with self-learnt adjacency matrix (dash line), CRM (dot-dash line), GNN without physics constraints (dot line). PI-GNN models are highlighted in blue.

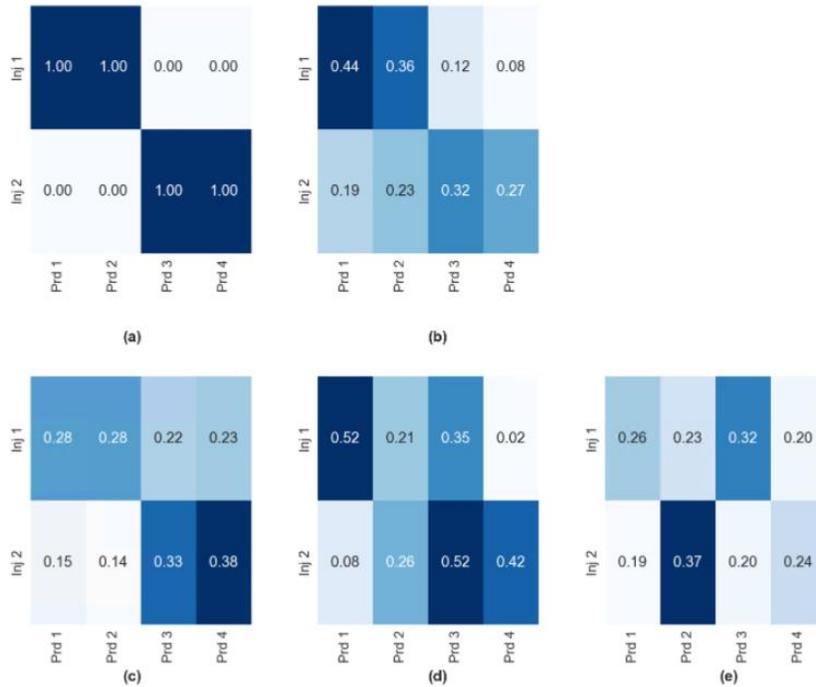

**Fig. 11** Case 4 (a) adjacency matrix informed by domain expertise (b) well connectivity matrix learnt from the PI-GNN model assisted with domain expertise (c) self-learnt well connectivity matrix with PI-GNN model (d) well connectivity matrix from CRM (e) well connectivity matrix leant from GNN baseline

| Method | Producer #1 | Producer #2 | Producer #3 | Producer #4 | Total |
|---|---|---|---|---|---|
| **Case 1** | | | | | |
| PI-GNN with domain knowledge assisted graph | **26.408** | **37.107** | 10.387 | 33.250 | **107.151** |
| PI-GNN with self-learnt graph | 76.295 | 83.342 | **8.600** | 60.224 | 228.460 |
| GNN baseline | 88.607 | 145.633 | 10.990 | **24.123** | 269.353 |
| CRM | 39.224 | 52.411 | 158.156 | 81.827 | 331.618 |
| **Case 2** | | | | | |
| PI-GNN with domain knowledge assisted graph | 88.344 | 90.395 | **68.826** | 40.450 | 288.013 |
| PI-GNN with self-learnt graph | **80.519** | **88.661** | 70.482 | **16.118** | **255.781** |
| GNN baseline | 121.968 | 126.816 | 178.881 | 31.302 | 458.966 |
| CRM | **57.898** | 153.025 | 111.507 | 525.146 | 847.575 |
| **Case 3** | | | | | |
| PI-GNN with domain knowledge assisted graph | 53.033 | 63.331 | 26.682 | 82.227 | **225.273** |
| PI-GNN with self-learnt graph | 122.070 | 120.834 | 6.471 | 50.220 | 299.594 |
| GNN baseline | 96.057 | 94.780 | **5.562** | 39.817 | 236.216 |
| CRM | **25.510** | **45.574** | 214.513 | 56.986 | 342.584 |
| **Case 4** | | | | | |
| PI-GNN with domain knowledge assisted graph | 85.070 | **32.922** | 27.827 | **65.398** | **211.216** |
| PI-GNN with self-learnt graph | 27.830 | 59.557 | 90.029 | 221.584 | 399.000 |
| GNN baseline | **26.069** | 70.350 | **22.819** | 179.208 | 298.445 |
| CRM | 92.543 | 84.148 | 274.616 | 97.308 | 548.614 |

**Table 1** RMSE comparisons. The lowest metric values for each case are highlighted in bold.

## Conclusion

We propose a physics-informed graph neural network method, PI-GNN, for subsurface production forecasting, which is computationally efficient, improves interpretability, and enhances the integration of domain expertise. Our proposed PI-GNN method outperforms the classic CRM and the GNN model without physics constraints as baselines for the demonstrated waterflood channelized reservoir model with randomized well location scenarios. An approximate estimation of the graph adjacency matrix based on domain expertise is a useful form of feature engineering that summarizes and integrates spatial information and assists model convergence for PI-GNN with a trainable adjacency matrix. For data-driven models, physics constraints are necessary for models to learn informative well connectivity. We design an innovative unsupervised training manner for PI-GNN in terms of time-dependent drainage volume and productivity index, which relaxes the assumptions of CRM. The PI-GNN model learns the behavior of drainage volume and productivity index without the requirement of corresponding measurements but by penalizing the violation of the mass balance term of the loss function. Nevertheless, the unsupervised predictions of drainage volume and productivity index, and the model performance can be potentially further improved with calibration for drainage volume and productivity index included as additional loss function terms.

## Acknowledgment

We would like to express our gratitude for the financial support from DIRECT consortium in the Hildebrand Department of Petroleum and Geosystem Engineering, The University of Texas at Austin.